\documentclass[11pt, a4paper, logo, copyright, nonumbering]{latex/mlins_template}
\usepackage[authoryear, sort&compress, round, comma]{natbib}
\setcitestyle{numbers,square}
\usepackage{dblfloatfix}
\usepackage{ulem}
\usepackage{caption}
\usepackage{dramatist}
\usepackage{xspace}
\usepackage{pifont} %
\usepackage{multirow}
\usepackage{tcolorbox}
\usepackage{xltabular}
\usepackage{longtable}
\usepackage{hyperref}
\usepackage{amsfonts}
\usepackage{amsmath}
\usepackage{amssymb}
\usepackage{lineno}
\usepackage{multirow}
\usepackage{adjustbox}

\usepackage[bottom]{footmisc}

\usepackage{CJKutf8}
\usepackage{subfigure}
\usepackage{setspace}

\usepackage{dsfont}
\usepackage{array} %
\usepackage{tabularx} %
\usepackage{subfigure} %
\usepackage{xcolor} %
\usepackage{tabularx}
\usepackage{booktabs}
\usepackage{algorithm}
\usepackage{listings}
\usepackage{parskip}
\usepackage{lipsum}  %
\usepackage{multicol} %
\usepackage{lineno}
\usepackage{wrapfig}

\makeatletter
\def\@BTrule[#1]{%
  \ifx\longtable\undefined
    \let\@BTswitch\@BTnormal
  \else\ifx\hline\LT@hline
    \nobreak
    \let\@BTswitch\@BLTrule
  \else
     \let\@BTswitch\@BTnormal
  \fi\fi
  \global\@thisrulewidth=#1\relax
  \ifnum\@thisruleclass=\tw@\vskip\@aboverulesep\else
  \ifnum\@lastruleclass=\z@\vskip\@aboverulesep\else
  \ifnum\@lastruleclass=\@ne\vskip\doublerulesep\fi\fi\fi
  \@BTswitch}
\makeatother

\addto\extrasenglish{
}

\renewcommand{\emph}[1]{\textit{#1}}

\renewcommand{\phi}{\varphi}

\renewcommand{\epsilon}{\varepsilon}
\renewcommand{\imath}{\mathrm{i}}

\newlength{\restsubwidth}
\newlength{\restsubheight}
\newlength{\restsubmoreheight}
\newcommand{\rest}[2]{%
        \settowidth{\restsubwidth}{\ensuremath{#2}}
        \settoheight{\restsubheight}{\ensuremath{{}_{#2}}}
        \ensuremath{{#1\hskip 0.5pt}_{\vrule\kern2pt\parbox[b][%
        4pt][b]{\the\restsubwidth}{%
                        \ensuremath{{}_{#2}}}}}
        }

\renewcommand{\today}{}

\title{\centering Intelligent Histology for Tumor Neurosurgery}

\author{
\centering
Xinhai Hou\textsuperscript{1},
Akhil Kondepudi\textsuperscript{1},
Cheng Jiang\textsuperscript{1},
Yiwei Lyu\textsuperscript{1},
Samir Harake\textsuperscript{1},
Asadur Chowdury\textsuperscript{1},
Anna-Katharina Mei\ss ner\textsuperscript{2},
Volker Neuschmelting\textsuperscript{2},
David Reinecke\textsuperscript{2},
Gina Furtjes\textsuperscript{2},
Georg Widhalm\textsuperscript{3},
Lisa Irina Koerner\textsuperscript{3},
Jakob Straehle\textsuperscript{4},
Nicolas Neidert\textsuperscript{4},
Pierre Scheffler\textsuperscript{4},
Juergen Beck\textsuperscript{4},
Michael Ivan\textsuperscript{5},
Ashish Shah\textsuperscript{5},
Aditya Pandey\textsuperscript{1},
Sandra Camelo-Piragua\textsuperscript{6},
Dieter Henrik Heiland\textsuperscript{7},
Oliver Schnell\textsuperscript{7},
Chris Freudiger\textsuperscript{8},
Jacob Young\textsuperscript{9},
Melike Pekmezci\textsuperscript{9},
Katie Scotford\textsuperscript{9},
Shawn Hervey-Jumper\textsuperscript{9},
Daniel Orringer\textsuperscript{10},
Mitchel Berger\textsuperscript{9},
Todd Hollon\textsuperscript{*}
}

\affil{
Department of Neurosurgery, University of Michigan, Ann Arbor, MI, USA\\
\textsuperscript{2}Department of Neurosurgery, University Hospital Cologne, Cologne, Germany\\
\textsuperscript{3}Department of Neurosurgery, Medical University of Vienna, Vienna, Austria\\
\textsuperscript{4}Department of Neurosurgery, University Medical Center Freiburg, Freiburg, Germany\\
\textsuperscript{5}Department of Neurosurgery, University of Miami, Miami, FL, USA\\
\textsuperscript{6}Department of Pathology, University of Michigan, Ann Arbor, MI, USA\\
\textsuperscript{7}Department of Neurosurgery, University Hospital Erlangen, Erlangen, Germany\\
\textsuperscript{8}Invenio Imaging Inc., Santa Clara, CA, USA\\
\textsuperscript{9}Department of Neurosurgery, University of California, San Francisco, CA, USA\\
\textsuperscript{10}Department of Neurosurgery, NYU Langone Health, New York, NY, USA\\

\textsuperscript{*}Corresponding Author
}

\begin{abstract}
The importance of rapid and accurate histologic analysis of surgical tissue in the operating room has been recognized for over a century. Our standard-of-care intraoperative pathology workflow is based on light microscopy and H\&E histology, which is slow, resource-intensive, and lacks real-time digital imaging capabilities. Here, we present an emerging and innovative method for intraoperative histologic analysis, called Intelligent Histology, that integrates artificial intelligence (AI) with stimulated Raman histology (SRH). SRH is a rapid, label-free, digital imaging method for real-time microscopic tumor tissue analysis. SRH generates high-resolution digital images of surgical specimens within seconds, enabling AI-driven tumor histologic analysis, molecular classification, and tumor infiltration detection. We review the scientific background, clinical translation, and future applications of intelligent histology in tumor neurosurgery. We focus on the major scientific and clinical studies that have demonstrated the transformative potential of intelligent histology across multiple neurosurgical specialties, including neurosurgical oncology, skull base, spine oncology, pediatric tumors, and periperal nerve tumors. Future directions include the development of AI foundation models through multi-institutional datasets, incorporating clinical and radiologic data for multimodal learning, and predicting patient outcomes. Intelligent histology represents a transformative intraoperative workflow that can reinvent real-time tumor analysis for 21\textsuperscript{st} century neurosurgery.

\end{abstract}

\begin{document}
\maketitle


\clearpage

\section*{Introduction}
The importance of rapid and accurate histologic analysis of surgical tissue in the operating room has been recognized for over a century \cite{Gal2005-na}. In 1930, Harvey Cushing described a technique for intraoperative histology of tumor specimens \cite{Eisenhardt1930-pe}. The method has evolved into modern intraoperative pathology, including cytologic preparations and frozen sectioning with hematoxylin and eosin (H\&E) staining \cite{Gal2005-na}. While light microscopy with H\&E staining is the standard for intraoperative pathologic anaylsis, it is labor and resource-intensive, limited to morphologic diagnoses (e.g. glial tumor), and does not scale well to serial or multiple specimens. These limitations have hindered histologic analysis from becoming the method of choice for tumor infiltration detection and surgical margin evaluation in neurosurgery. Moreover, because intraoperative H\&E histology does not produce real-time \emph{digital} images, integrating artificial intelligence (AI) for diagnostic decision support has been infeasible.

\begin{figure}
    \centering
    \includegraphics[width=1\linewidth]{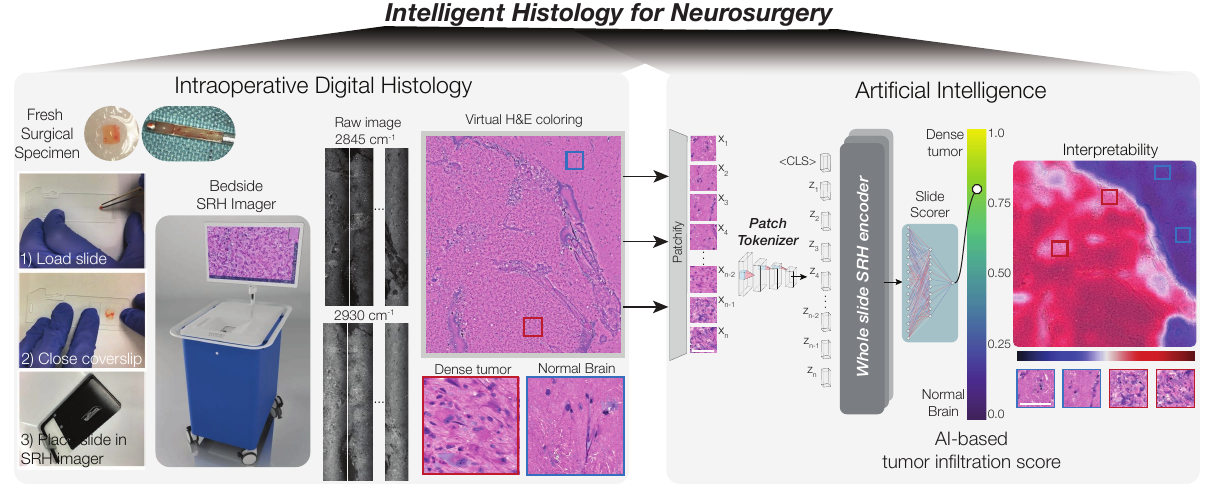}
    \caption{\small \textbf{Overview of Intelligent Histology}. A fresh, unprocessed neurosurgical specimen is obtained via an open or stereotactic biopsy. The tissue is then loaded into a premade microscope slide, the coverslip is closed, and the slide is placed in the SRH imager (FDA-approved NIO Imaging System, Invenio Imaging, Inc.). Raw SRH images are acquired as strips at two Raman shifts, 2,845 cm\textsuperscript{$-1$} (lipid channel) and 2,930 cm\textsuperscript{$-1$} (protein channel). The time to acquire a 3X3-mm2 SRH image is approximately 90 seconds. Raw images are colored using a virtual H\&E colorscheme for clinician review. AI models can be trained using state-of-the-art deep learning methods for tumor infiltration detection and quantification, for example \cite{Kondepudi2025-ll}. Our intelligent histology models are interpretable because, in addition to providing a whole-slide prediction, region-level attention or prediction offers segmentation of the most diagnostic, tumor-infiltrated regions within an SRH image.}
    \label{fig:fig_1_ihisto}
\end{figure}

A major open problem for 21\textsuperscript{st} century neurosurgery is developing innovative methods to improve the intraoperative analysis of surgical tissue. The United States has identified this topic as a major area of innovation for precision healthcare \cite{noauthor_undated-zr}. Here, we review an emerging method, which we call \emph{Intelligent Histology}, that combines AI and stimulated Raman histology (SRH), a label-free optical imaging method, for real-time microscopic specimen analysis, tissue diagnosis, and tumor infiltration detection (Figure \ref{fig:fig_1_ihisto}). We review the scientific background, clinical translation, and future directions of intelligent histology. Combining state-of-the-art AI and fast optical imaging is an innovative step towards developing a 21\textsuperscript{st} century intraoperative pathology workflow for neurosurgery.

\section*{Background}

\subsection*{Label-free Imaging with SRH}
\begin{figure}[t]
    \centering
    \includegraphics[width=1\linewidth]{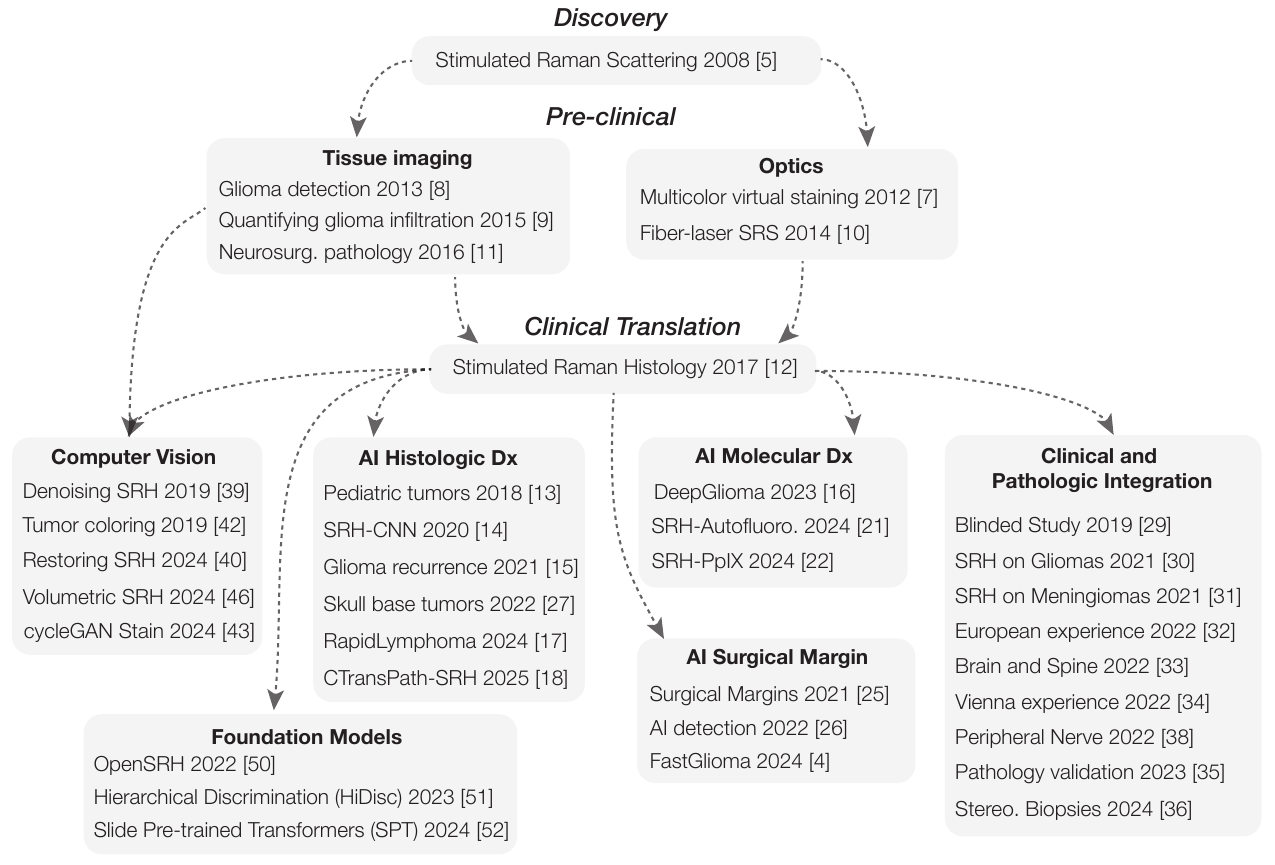}
    \caption{\textbf{Intelligent Histology Timeline}.}
    \label{fig:fig_2_timeline}
\end{figure}

Stimulated Raman scattering microscopy was first described in 2008 by Freudiger et al. \cite{Freudiger2008-gj} (Figure \ref{fig:fig_2_timeline}). Stimulated Raman scattering is a vibrational microscopy based on Raman scattering that generates high-resolution images \emph{without} requiring fluorescent labels or dyes. A major advantage of stimulated Raman over other optical imaging methods is that the intensity of the Raman signal is directly proportional to the concentration of macromolecules within biological tissue. The differences in biochemical concentrations of lipids, proteins, and nucleic acids within the tissue generate label-free image contrast. While spontaneous Raman spectroscopy has been used in neurosurgery, stimulated Raman has a 1000X greater signal-to-noise ratio for image acquisition and tissue analysis \cite{Hollon2016-uk}. The increased signal-to-noise ratio of stimulated Raman has facilitated clinical translation and adoption.

\begin{figure}
    \includegraphics[width=1\linewidth]{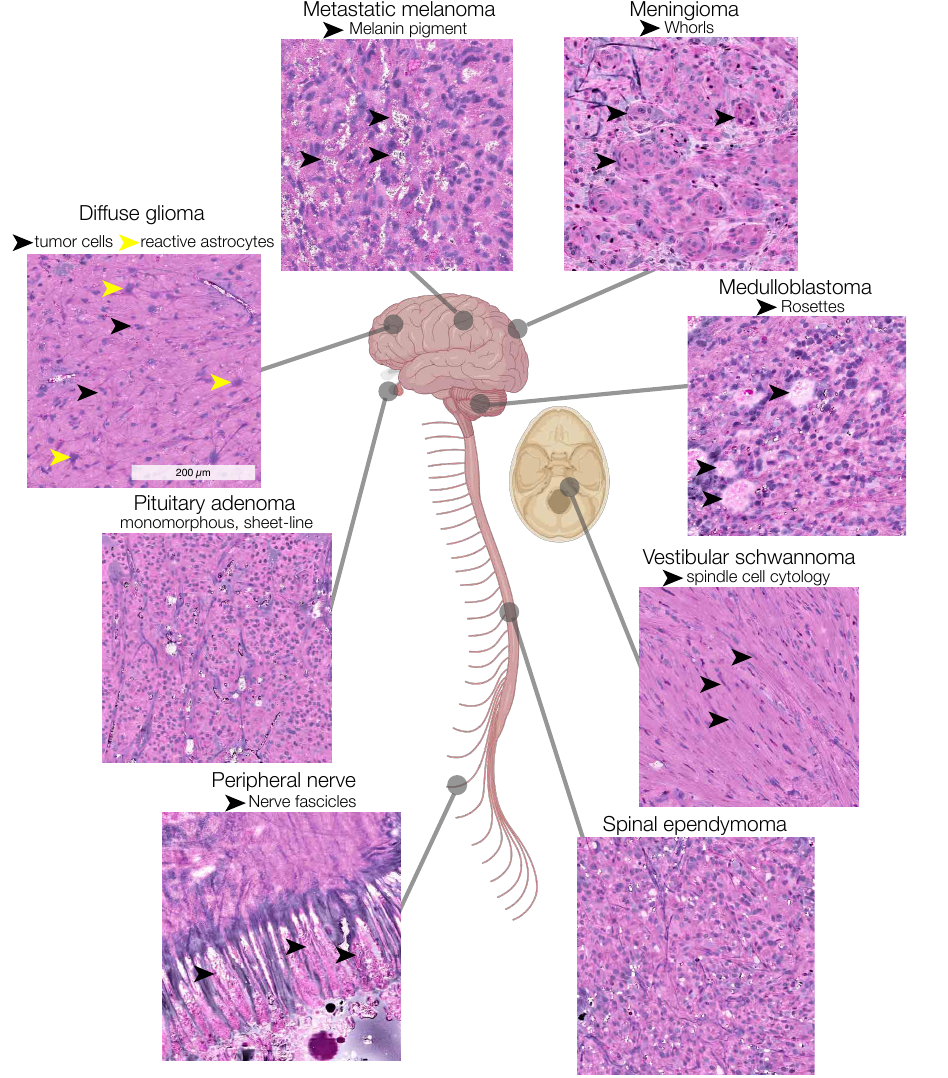}
    \caption{\small \textbf{Comprehensive Digital Histology for Tumor Neurosurgery}. A diverse panel of tumors imaged using rapid, intraoperative SRH. A virtual H\&E color scheme is applied for easier intraoperative interpretation for surgeons and pathologists. Pathognomonic features, such as melanin pigment, meningeal whorls, and Homer Wright rosettes, are seen in SRH images, enabling fast and accurate histologic diagnosis for both clinicians and AI models.}
    \label{fig:fig_3_panel}
\end{figure}

After a sequence of preclinical breakthroughs between 2012 and 2016 \cite{Freudiger2012-gv, Ji2013-zw, Ji2015-uz, Freudiger2014-at, Lu2016-he}, Orringer et al. presented the definitive clinical translation of SRH \cite{Orringer2017-nn}. The clinical SRH imager was a portable fiber-laser-based stimulated Raman scattering microscope used at the bedside to image fresh, unprocessed surgical specimens within minutes of tissue biopsy. SRH captures high-resolution, digital images that allow for real-time histologic analysis by pathologists and surgeons. The authors developed a virtual H\&E staining method for coloring two-channel, greyscale SRH images that can be used for intraoperative pathology (Figure \ref{fig:fig_3_panel}). For intraoperative tumor classification, the pathologists' diagnoses were concordant when using SRH versus standard H\&E pathology. These results demonstrated the feasibility of SRH workflow for tissue analysis and diagnosis.

\subsection*{The Beginnings of AI and SRH}
SRH is an ideal imaging method for AI because it generates high-resolution, quantitative, and digital images for machine learning. The first application of machine learning to SRH used a linear model to quantify degrees of tumor infiltration based on cellularity, axonal density, and SRH channel intensity ratios \cite{Ji2015-uz}. Subsequent early work used multilayer perceptrons and random forest models on hand-engineered SRH features for coarse tumor diagnoses in adult and pediatric brain tumors \cite{Orringer2017-nn, Hollon2018-hl}. These early applications of machine learning to SRH predate the widespread use of deep learning and the recent explosion of AI foundation models. However, they showed the feasibility of training large-scale SRH-AI models and provided the groundwork for developing intelligent histology.

\section*{Clinical Translation of Intelligent Histology}
\subsection*{Histologic diagnosis of central nervous system tumors}
The major role of intraoperative pathology is to analyze surgical tissue and provide a timely histology diagnosis to inform surgical decision-making. A study in 2020 used a convolutional neural network (CNN), trained on over 2.5 million SRH images, to classify the 13 most common histologic diagnoses in neurosurgical oncology \cite{Hollon2020-ez}. In a prospective clinical trial, the SRH-CNN achieved a diagnostic accuracy of 93.1\%, equivalent to pathologists with standard H\&E histology, but an order of magnitude faster (2 minutes versus 20 minutes). The CNN model learned to identify interpretable histologic features in SRH images, such as hypercellularity, pleomorphism, chromatin structure, and axonal density. Moreover, the SRH-CNN model discovered known histologic features associated with tumor malignancy and progression, such as anaplasia and high nuclei-cytoplasm ratios, and used these to differentiate low grade and high grade gliomas. A follow-up study fine-tuned the original SRH-CNN model to differentiate tumor recurrence from treatment effect/pseudoprogression \cite{Hollon2020-cl}. This task is challenging for intraoperative pathology because previous treatment can induce histologic changes that mimic tumor recurrence, such as gliosis, inflammation, and necrosis. The model achieved a classification accuracy of 95.8\% on an external testing cohort of recurrent diffuse gliomas patients.

\begin{figure}[t]
    \centering
    \includegraphics[scale=0.8]{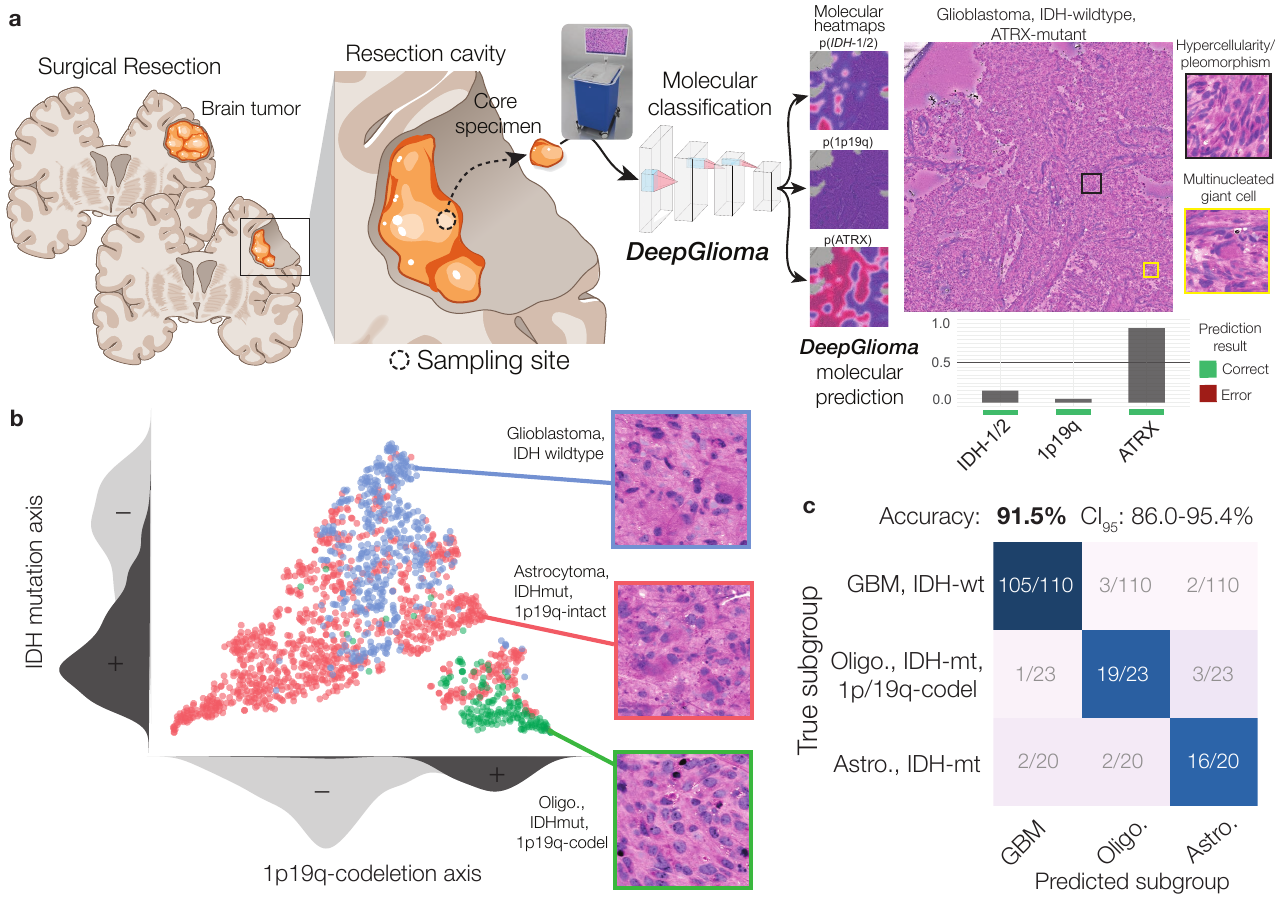}
    \caption{\small \textbf{Molecular Classification of Diffuse Gliomas with Intelligent Histology}. \textbf{a}, A patient with a diffuse gliomas undergoes surgical resection. A core specimen is collected and imaged intraoperative using the SRH imager. \emph{\textbf{DeepGlioma}} then performs real-time, automated molecular marker prediction, including IDH mutation, 1p19q co-deletion, and ATRX mutation, to achieve a final WHO classification \emph{within 2 minutes of biopsy}. \textbf{b}, tSNE plot of DeepGlioma outputs shows that the model can differentiate the three major diffuse gliomas subtypes by separating them on an IDH axis and 1p19q axis. \textbf{c}, DeepGlioma classification accuracy surpasses 90\% on a prospective, multicenter testing cohort of diffuse glioma patients \cite{Hollon2023-uf}.}
    \label{fig:fig_4_molecular}
\end{figure}

Differentiating surgical versus non-surgical tumors intraoperatively is essential to define surgical goals. Reinecke et al. developed RapidLymphoma to detect and classify primary central nervous system lymphomas using intelligent histology \cite{Reinecke2024-jm}. Because lymphomas are less common than other central nervous system tumors, a major innovation of RapidLymphoma was using self-supervised pretraining, similar to generative pre-trained transformer (GPT). In a prospective, multi-center testing cohort, RapidLymphoma achieved a overall balanced accuracy of 97.8\% for detecting primary CNS lymphoma, outperforming frozen sectioning performance. A related innovative work from Scheffle et al. used an open-source computational pathology model, CTransPath, for automated SRH feature extraction and lymphoma classification \cite{Scheffler2025-qt}. To our knowledge, this is the first instance of integrating an off-the-shelf computational pathology model for SRH fine-tuning. Importantly, they achieved comparable classification as RapidLymphoma using a significantly smaller training dataset.

\begin{figure}[t]
    \centering
    \includegraphics[scale=0.67]{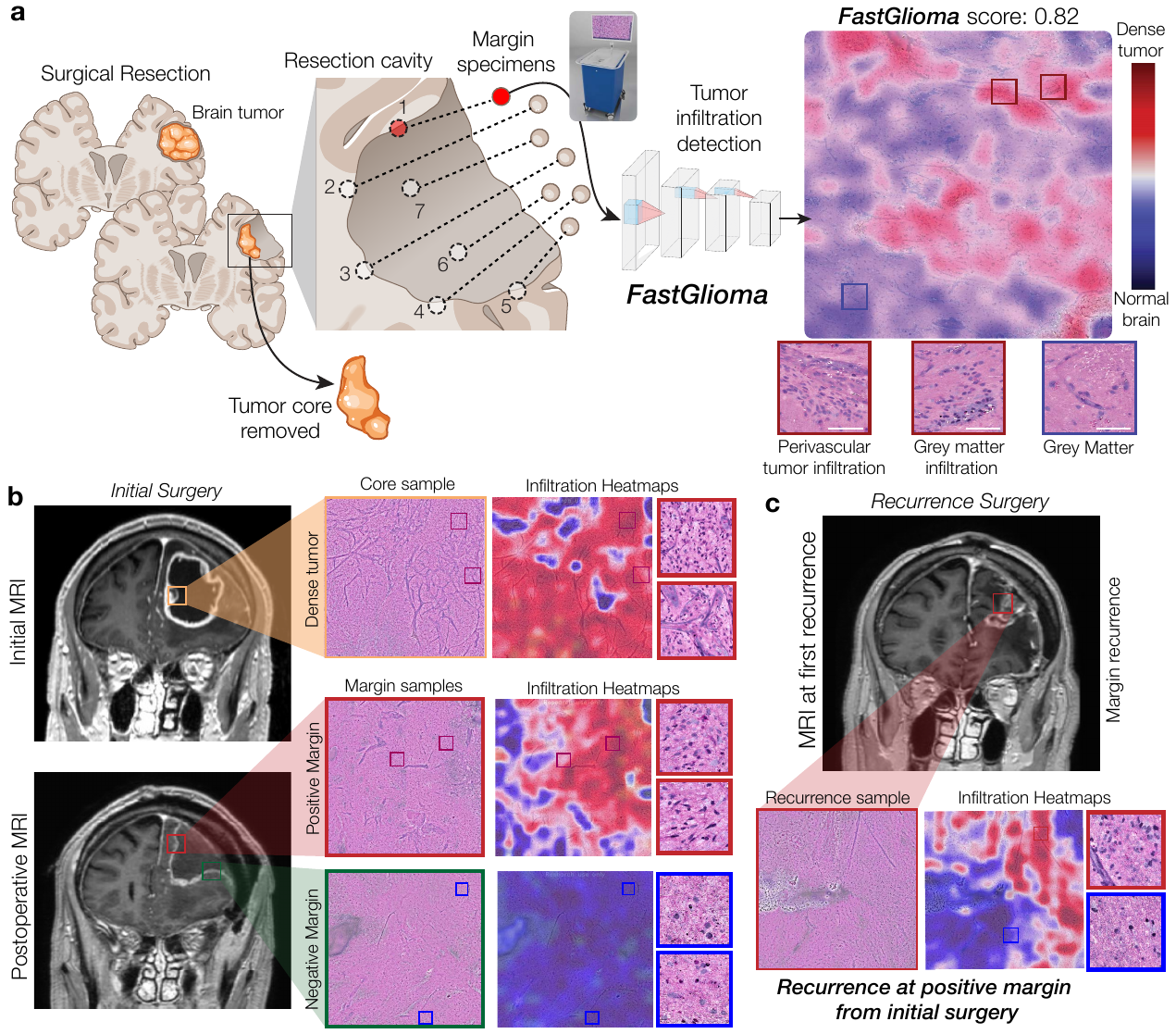}
    \caption{\small \textbf{Tumor Infiltration Detection with Intelligent Histology} \textbf{a}, A patient with a brain tumor undergoes surgical resection. Following resection of the tumor core, margin specimens are collected for SRH imaging and AI-based detection of tumor infiltration. \textbf{\emph{FastGlioma}} outputs a heatmap that identifies regions of tumor infiltration and a normalized score between 0 (normal) and 1 (dense tumor). \textbf{b}, A illustrative case a patient who underwent \emph{FastGlioma}-guided tumor resection at their initial surgery. A positive margin was identified at the medial border of the resection cavity. \textbf{c}, Patient subsequently presented with a recurrence along the medial resection cavity at the positive margin site. These results demonstrate that \emph{FastGlioma} can detect microscopic tumor burden and guide surgical resections, potentially prolonging progression-free and overall survival.}
    \label{fig:fig_5_margin}
\end{figure}

\subsection*{Molecular classification of brain tumors}
Molecular genetics are now a major factor in classifying central nervous system tumors \cite{Louis2021-vd}. Adult-type diffuse gliomas are classified by isocitrate dehydrogenase-1/2 (IDH) mutations and 1p19q co-deletions (1p19q). Importantly, Hervey-Jumper et al. demonstrated that surgical goals, such as gross total versus supratotal resection, should be informed by IDH and 1p19q status \cite{Hervey-Jumper2023-mi}. Unfortunately, conventional intraoperative pathology does not provide molecular data. Intelligent histology has the potential to predict molecular markers from SRH images to inform surgical goals. DeepGlioma was developed in 2023 for rapid and accurate molecular classification of diffuse gliomas \cite{Hollon2023-uf} (Figure \ref{fig:fig_4_molecular}). DeepGlioma was the first multimodal intelligent histology model trained with SRH and large-scale, public genomic data. DeepGlioma was first trained to learn the genomic landscape of diffuse gliomas, such as co-occurence statistics of IDH mutations and 1p19q co-deletions. Then, the model was trained to use this landscape to classify gliomas based on SRH image features. In a prospective, multicenter, international testing cohort of patients with diffuse glioma (n=153), DeepGlioma acheived a mean accuracy of 93.3±1.6\% for predicting the molecular markers used by the World Health Organization to define diffuse gliomas, including IDH mutation, 1p19q co-deletion, and ATRX mutation.

Intelligent histology has characterized two-photon fluorescence in surgical specimens \cite{Furtjes2023-pz, Nasir-Moin2024-af}. The SRH-CNN model \cite{Hollon2020-ez} was used to identify tumor infiltrated regions in a diverse set of surgical specimens. Spatially registered two-photon fluorescence images were acquired and the mean autofluorescence was measured within AI-selected tumor infiltrated regions. They found autofluorescence in the brain varies depending on the tissue type and localization and differs significantly among various brain tumors. A subsequent study elucidates the molecular and spatial relationship between SRH image features and protoporphyrin IX (PpIX) fluorescence due to 5-aminolevulinic acid (5-ALA) used in glioma surgery \cite{Nasir-Moin2024-af}. Using 115 high-grade glioma patients from four medical centers, the authors discovered five distinct patterns of PpIX fluorescence using semi-supervised learning. Spatial transcriptomic analyses of the imaged tissue demonstrated that myeloid cells predominate in areas where PpIX accumulates in the intracellular space. Additional analysis of spatially resolved metabolomics, transcriptomics and RNA-sequencing data confirmed that myeloid cells preferentially accumulate and metabolize PpIX. These findings demonstrate how intelligent histology can be used to shed new light on existing surgical tools and the immune microenviroment of gliomas.

\subsection*{Surgical margin analysis}
Safe maximal resection has stood the test of time as an essential component in the management of brain tumors \cite{Hervey-Jumper2023-za}. Residual tumor burden after surgery is the major risk factor for a worse overall prognosis in the majority of brain tumor diagnoses. Unfortunately, clinical studies have demonstrated that many patients, perhaps the majority, do not receive optimal surgical treatment  \cite{Orringer2012-fb}. Dense, safely resectable tumor infiltration is found at the surgical margin in 30\% of diffuse gliomas patients \cite{Pekmezci2021-et}. Developing on previous work, Reinecke et al. showed that histologic classification models can be leveraged for tumor infiltration detection and validated this approach in a large external cohort \cite{Reinecke2022-eo}.

FastGlioma is an intelligent histology model for fast (<10 seconds) and accurate detection of glioma infiltration at the surgical margin \cite{Kondepudi2025-ll} (Figure \ref{fig:fig_5_margin}). FastGlioma was pretrained on a large-scale SRH dataset, around 4 million images, using self-supervision, that is without labels. The model was then fined-tuned on a small, expert-annotated tumor infiltration dataset to output a tumor infiltration score between 0 and 1, 0 meaning normal brain and 1 meaning dense tumor \cite{Pekmezci2021-et}. In a prospective, multicentre, international testing cohort of patients with diffuse glioma (n=220), FastGlioma was able to detect and quantify the degree of tumour infiltration with an average area under the receiver operating characteristic curve of 92.1 +/- 0.9\%. FastGlioma outperformed image-guided and fluorescence-guided adjuncts for detecting tumour infiltration during surgery by a wide margin in a head-to-head, prospective study (n=129). We encountered no adverse events with surgical margin sampling. FastGlioma was performant across all WHO diffuse glioma molecular subtypes. Additionally, FastGlioma showed generalization to other adult and paediatric brain tumour diagnoses, such as metastatic tumors and diffuse midline gliomas, demonstrating the potential of FastGlioma as a general-purpose adjunct for guiding brain tumor surgeries.

The clinical importance of surgical margin analysis in not limited to diffuse gliomas. Residual tumor burden after meningioma or pituitary surgery, for example, is known to be a strong predictor of future recurrence. Jiang et al. showed that microscopic tumor infiltration can be detected using intelligent histology in grossly normal dura at the resection margin of meningiomas \cite{Jiang2022-ea}. In pituitary adenoma surgery, he showed that sub-millimeter ACTH-secreting pituitary adenomas can be detected in intraoperative pituitary specimens. Skull base surgeons have highlighted the potential of intelligent histology in sampling the medial wall of the cavernous sinus for adenoma invasion during pituitary surgery to offer real-time feedback on surgical decision-making \cite{Mohyeldin2022-ak}.

\subsection*{Integrating intelligent histology into neurosurgery}
A global effort has emerged towards integrating SRH and intelligent histology into neurosurgical practice. Multiple independent groups have shown the benefits of using SRH in the operating room to rapidly evaluate surgical specimens. In a prospective blinded study, University of Miami demonstrated that SRH-based diagnosis was on average 30 minutes faster than frozen section-based diagnoses, with near perfect diagnostic concordance (Cophen's $\kappa$ = 0.834, p < 0.0001) between SRH and permanent/frozen sections \cite{Eichberg2019-rd}. They subsequently published their clinical SRH experience in meningiomas and gliomas \cite{Di2021-kg, Di2021-wk}. Paired publications from the University of Freiburg presented their experience integrating SRH into a large European neurosurgical center \cite{Neidert2022-ku, Straehle2022-on}. Notably, they presented their experience with spine tumors, showing comparable diagnostic accuracy and SRH/H\&E concordance with brain tumors. The University of Vienna presented their experience with intraoperative SRH across a wide range of brain tumors and non-neoplastic lesions, such as cavernomas and epidermoid cysts. Imporantly, they showed excellent concordance between SRH and H\&E to assess degrees of cellularity and tumor infiltration at the surgical margin \cite{Wadiura2022-qd}. 

Movahed-Ezazia et al. performed a 1 year study at New York University to evaluate the median turnaround time (TAT) of conventional neuropathology frozen sections with prospective SRH. The median TAT for SRH diagnosis was 3.7 minutes, approximately 10X faster than the median frozen section TAT, 31 minutes \cite{Movahed-Ezazi2023-gw}. This supports the feasibility and time-savings of implementing SRH as a rapid method for intraoperative neuropathology. Reinecke et al. presented a comprehensive study from the University of Cologne on using intelligent histology for stereotactic brain biopsies, incorporating both histologic and molecular classification models \cite{Reinecke2024-cs, Hollon2020-ez, Hollon2023-uf}. Firstly, they demonstrated that intelligent histology yields accurate predictions on stereotactic brain biopsies. Critically, they showed that biopsy sample size effects diagnostic accuracy and systematically found a minimal size/surface area required for accurate diagnosis. This was the first study to evaluate the relationship between tissue specimen size and model performance. Mei\ss ner et al. then showed that frozen tissue samples from a tissue biobank can be used to generate high-quality images for AI training, thereby increasing data examples of rare tumors and diagnoses \cite{Meissner2024-qt}.

Finally, intelligent histology has been applied to peripheral nerve tumor surgery \cite{Wilson2022-wb}. SRH effectively visualizes lipids and may be well-suited for intraoperative assessment of peripheral nerve quality. Wilson et al. used SRH intraoperatively on six peripheral nerve cases to generate high-resolution images of myelinated axons within nerve fascicles. The results suggest that SRH provides a reliable and quantitative method for evaluating myelin in peripheral nerves, potentially improving surgical decision-making for nerve repair.

\begin{figure}
    \vspace{-20pt}
    \centering
    \includegraphics[width=1\linewidth]{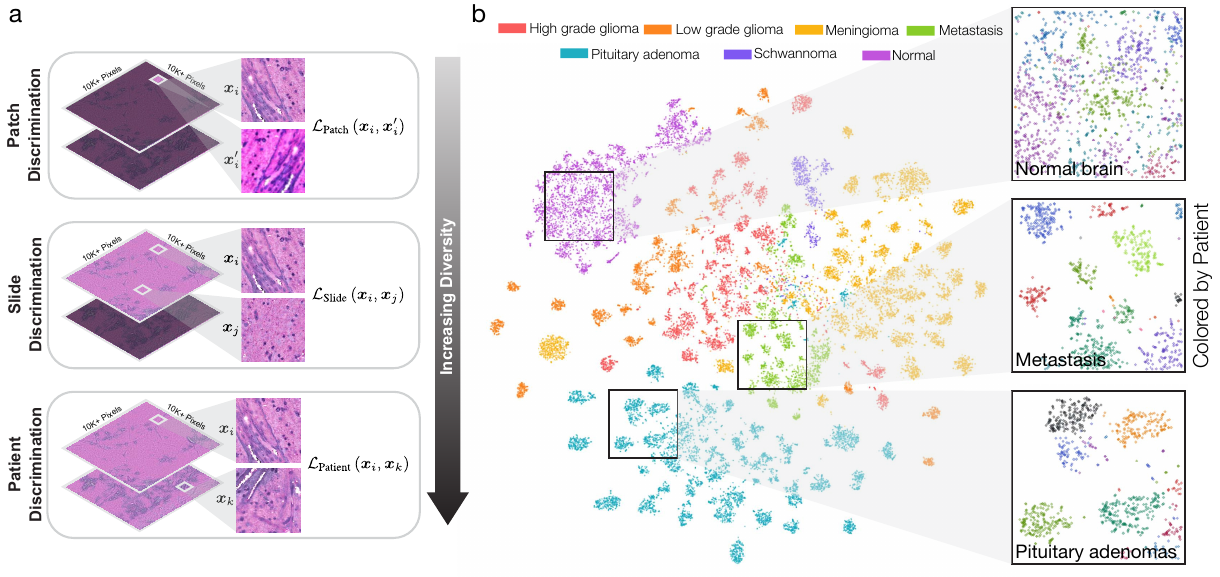}
    \caption{\small \textbf{Foundation Models for Intelligent Histology}. \textbf{a}, Clinical biomedical microscopy and SRH has a hierarchical patch-slide-patient data structure. This structured can be used to train foundation models without the need to data labels. Hierarchical discrimination (HiDisc) combines patch, slide, and patient discrimination into a unified self-supervised learning task that can be used to train intelligent histology foundation models. \textbf{b}, After training with HiDisc, we show the tSNE plot of SRH images with the diagnosis labels. Note that the model was \emph{not} trained with labels, yet learns the differentiate the major tumor types in neurosurgery. Moreover, it captures diversity and heterogeneity for each specimen (right). The model captures unique features for each patient, demonstrated by the clustering of regions from the same patient.}
    \label{fig:fig_6_foundation}
    \vspace{-10pt}
\end{figure}


\subsection*{Biomedical computer vision}
Intelligent histology has driven new developments in AI-based biomedical microscopy and computer vision. A major area has been improving SRH image quality through denoising, restoration, colorization, and super-resolution. Manifold et al. trained a U-Net model to denoise SRH images acquired at low laser power \cite{Manifold2019-ft}. Subsequent work developed a generative SRH image denoising and restoration method using diffusion models called restorative step-calibrated diffusion (RSCD) \cite{Lyu2024-bo}. RSCD can take any SRH image, with unknown sources or severity of image degradation, and restore the image to diagnostic quality for pathologic review and AI prediction. With advances in image-to-image models \cite{Zhu2017-wf}, virtual coloring of SRH images has improved. Shin et al. showed that tailored coloring, beyond virtual H\&E staining, that utilizes rich SRH chemical information can improve diagnostic accuracy of skull base tumors \cite{Shin2019-pl}. In 2024, Liu at al. trained a coloring generative adversarial network to convert greyscale SRH images to an H\&E coloring perceptually indistinguishable from laboratory H\&E staining \cite{Liu2024-ja}. The role of 3D microscopy in biomedical research and computational pathology is growing \cite{Xie2022-gs, Song2024-hu}. Jiang et al. presented the first intelligence histology model to generate high-resolution 3D SRH images \cite{Jiang2024-wv}. Using a generative diffusion model, 2D SRH images are converted into 3D images that better capture cellular morphology, vascularity, and tumor infiltration. 


\section*{Future Directions of Intelligent Histology}

\subsection*{Foundation Models for Intelligent Histology}
Foundations models are any AI model trained on broad, diverse, and minimally uncurated data, generally using self-supervision (i.e. no labels), that can be adapted or fine-tuned to a wide range of downstream tasks. Foundation modeling is the leading candidate for developing generalist medical AI \cite{Moor2023-av}. A major future direction of intelligent histology is developing foundation models to solve multiple diagnostic or prognostic tasks. Previous work has trained models for a single task on a curated dataset, such as histologic diagnosis or tumor infiltration quantification. Foundation model research has shown that training models with larger datasets, followed by fine-tuning, results in better generalization and performance \cite{Kaplan2020-ia, Brown2020-dv}. These results motivated the large-scale pre-training of FastGlioma before tumor infiltration fine-tuning \cite{Kondepudi2025-ll}.

To promote open science and the development of foundation models for intelligent histology, we released \href{https://opensrh.mlins.org/}{OpenSRH}, the first public dataset of clinical SRH images from 300+ brain tumors patients and 1300+ unique whole slide SRH images \cite{Jiang2022-cj}. OpenSRH contains data from the most common brain tumors diagnoses, full pathologic annotations, whole slide tumor segmentations, raw and processed. We hope to foster mutli-institutional datasets that will serve as the basis for training foundation models. Incorporating OpenSRH, Jiang, Hou, et al. published two strategies for foundation model training that address SRH region-level training, called HiDisc \cite{Jiang2023-rq} and whole slide-level training, called Slide Pre-trained Transformers (SPT) \cite{Hou2024-bj} (Figure \ref{fig:fig_6_foundation}). Both HiDisc and SPT scale to large datasets and use self-supervision only to train intelligent histology models. We aim to develop multi-institutional consortia that will facilitate SRH data sharing and foundation model development. 

\subsection*{Multimodal learning}
Multimodal models can take as input data from two or more domains. Multimodal models, such as DALL-E \cite{Ramesh2022-bv} and LLaVa \cite{Liu2023-ey}, can take images or text as input and output images or text. The major advantage of multimodal learning in medical AI is that most diagnostic or prognostic tasks should include text, tabular, and image data to improve clinical context. Pathologist and radiologist use clinical context when providing pathology or radiology reports. Recent computational pathology models have embraced multimodal learning to improve performance across multiple cancer types and include visual question answering \cite{Lu2024-be}. For example, interpreting SRH images of brain tumors will be aided by knowing the patient's age, presenting symptoms, tumor location, and MRI features. Including these additional variables for clinical context will improve overall performance and prevent diagnostic errors. Future intelligent histology models will benefit from being multimodal to better replicate how clinical decisions incorporate information from multiple sources.

\subsection*{Patient outcome prediction}
Predicting patient outcomes is arguably the most important AI task in precision medicine. Prognostication, or forecasting future events from past and present data, is one of the more challenging machine learning tasks due to limited data and unforseen or random events. Recent computation pathology research has studied stratifying patient cancer survival from H\&E images and genetic data \cite{Chen2022-gv, Wang2024-tn}. While this work represents an important step forward, this strategy will always produce suboptimal results due to not accounting for important prognostic information, such as preoperative tumor volumes, exent of resection, functional status, age, etc. This re-emphasizes the essential role of multimodal models for patient outcome prediction. Current multimodal intelligent histology research is underway to incorporate patient demographic, clinical, radiologic, surgical, and pathologic information to stratify survival for diffuse glioma patients.



\section*{Conclusions}
Intelligent histology unifies rapid digital histology and deep learning to advance intraoperative neuropathology. Neurosurgeons now have access to timely and reliable AI-based interpretation of tumor type, molecular profile, and margin status at the patient's bedside, enabling safer, more complete resections and informed surgical decision-making. Successive intelligent histology breakthroughs chart a clear trajectory toward more comprehensive and interpretable support systems that generalize across diverse tumor types and operative workflows. Open-source resources like OpenSRH, together with early self-supervised foundation models, signal a collaborative shift toward large-scale, multimodal, multicenter AI development and validation. Collectively, these advances demonstrate a transformative intraoperative pathology paradigm in which every fresh specimen becomes a rich digital substrate for AI, accelerating diagnosis, guiding precision surgery, and deepening our understanding of tumor biology for the benefit of patients and our neurosurgical community.

\bibliographystyle{unsrt}
\bibliography{latex/paperpile}

\end{document}